\documentclass[runningheads]{llncs}

\usepackage{graphicx}
\usepackage{epstopdf}
\usepackage{algorithm}
\usepackage{algorithmic}
\usepackage{setspace}
\usepackage{float}
\usepackage{array}
\usepackage{booktabs}
\usepackage{threeparttable}
\usepackage{cite}
\usepackage{color,array}
\usepackage{multirow}
\usepackage{makecell}
\usepackage{mathtools}
\usepackage{amsmath,amssymb}

\usepackage{graphicx}

\newcommand{\tabincell}[2]{\begin{tabular}{@{}#1@{}}#2\end{tabular}}

\begin{document}

\title{Imperceptible and Multi-channel Backdoor Attack against Deep Neural Networks}

\titlerunning{Imperceptible and Multi-channel Backdoor Attack against DNN}

\author{Mingfu Xue\inst{1} \and
Shifeng Ni \inst{1} \and
Yinghao Wu \inst{1} \and
Yushu Zhang \inst{1} \and
Jian Wang \inst{1} \and
Weiqiang Liu \inst{2}}

\authorrunning{M. Xue et al.}

\institute{College of Computer Science and Technology, Nanjing University of Aeronautics and Astronautics, Nanjing, China\\
\and
College of Electronic and Information Engineering, Nanjing University of Aeronautics and Astronautics, Nanjing, China\\}

\maketitle              

\begin{abstract}
Recent researches demonstrate that Deep Neural Networks (DNN) models are vulnerable to backdoor attacks.
The backdoored DNN model will behave maliciously when images containing backdoor triggers arrive.
To date, existing backdoor attacks are single-trigger and single-target attacks, and the triggers of most existing backdoor attacks are obvious thus are easy to be detected or noticed.
In this paper, we propose a novel imperceptible and multi-channel backdoor attack against Deep Neural Networks by exploiting Discrete Cosine Transform (DCT) steganography.
Based on the proposed backdoor attack method, we implement two variants of backdoor attacks, i.e., N-to-N backdoor attack and N-to-One backdoor attack.
Specifically, for a colored image, we utilize DCT steganography to construct the trigger on different channels of the image. As a result, the trigger is stealthy and natural.
Based on the proposed method, we implement multi-target and multi-trigger backdoor attacks.
Experimental results demonstrate that the average attack success rate of the N-to-N backdoor attack is 93.95\% on CIFAR-10 dataset and 91.55\% on TinyImageNet dataset, respectively.
The average attack success rate of N-to-One attack is 90.22\% and 89.53\% on CIFAR-10 and TinyImageNet datasets, respectively.
Meanwhile, the proposed backdoor attack does not affect the classification accuracy of the DNN model.
Moreover, the proposed attack is demonstrated to be robust to the state-of-the-art backdoor defense (Neural Cleanse).

\end{abstract}

\section{Introduction}
Deep Neural Networks (DNN) model has made significant achievements in various fields, such as image classification \cite{he2016deep}, speech recognition \cite{amodei2016deep}, automatic driving \cite{chen2015deepdriving}, and so on.
However, recent studies have shown that DNN is vulnerable to backdoor attacks.
In the backdoor attacks, an attacker injects images pasted with trigger into the training set and changes their labels to the target label, then the attacker uses the backdoor training set to train the model.
The model trained on the backdoor training set will show a normal classification performance on clean images.
However, when backdoor instances arrive, the backdoored model will output the target label.

To date, a number of backdoor attacks have been proposed, which face the following challenges:
(i) The triggers of most existing backdoor attacks are obvious patterns on the images, thus are easy to be noticed by humans.
(ii) Almost all the existing backdoor attacks \cite{abs-1708-06733,0005LMZ21,ZhongLSZ020,ning2021invisible,LiXZZZ21,LiuMALZW018} are single-trigger and single-target backdoor attacks, called One-to-One attack in this paper.
So far, there has been only one work on One-to-N and N-to-One backdoor attack \cite{xue2020one}, however, the trigger in \cite{xue2020one} is obvious.
(iii) State-of-the-art backdoor defense methods can detect most of the existing backdoor attacks.

In this paper, we propose an imperceptible and multi-channel backdoor attack against DNN by exploiting Discrete Cosine Transform (DCT) steganography \cite{taburet2020jpeg, mohd2000implementation}.
Further, based on the proposed imperceptible and multi-channel backdoor attack method, we implement two variants of backdoor attacks, i.e., imperceptible N-to-N backdoor attack and imperceptible N-to-One backdoor attack.
Specifically, we utilize DCT steganography to embedded multiple triggers into different channels (i.e., RGB Channels) of an image.
For the proposed imperceptible N-to-N attack, the triggers embedded into different channels of the image can activate different backdoor targets respectively.
For the proposed imperceptible N-to-One attack, the backdoor can only be activated when all the triggers embedded in different channels of the image are present simultaneously, thus is very stealthy.

Experimental results demonstrate that the average success rate of N-to-One backdoor attack is 90.22\% and 89.53\% on CIFAR-10 \cite{Krizhevsky_2009_17719} and TinyImageNet \cite{le2015tiny} datasets, respectively.
For N-to-N backdoor attack ($N = 3$), when the triggers are embedded into the Red, Green and Blue channel of the image, the average success rate is 93.04\% (Red), 95.09\% (Green), 93.95\% (Blue) on CIFAR-10 \cite{Krizhevsky_2009_17719} dataset and 93.29\% (Red), 93.09\% (Green), 91.55\% (Blue) on TinyImageNet \cite{le2015tiny} dataset, respectively.

The main contributions of this paper are as follows:
\begin{enumerate}
\item{To the best of our knowledge, this paper proposes the first imperceptible and multi-channel backdoor attack against Deep Neural Networks.}
\item{Based on our proposed backdoor attack method, we propose/implement two variants of backdoor attacks, N-to-N backdoor attack and N-to-One backdoor attack, while almost all the existing works are One-to-One attacks \cite{abs-1708-06733,0005LMZ21,ZhongLSZ020,ning2021invisible,LiXZZZ21,LiuMALZW018}.}
\item{The proposed backdoor attack is demonstrated to be robust to the state-of-the-art backdoor defense (Neural Cleanse \cite{wang2019neural}).}
\end{enumerate}

The rest of this paper is organized as follows.
The preliminary and related work are reviewed in Section \ref{sec:sec2}.
The proposed backdoor attack is elaborated in Section \ref{sec:sec3}.
The experimental results are discussed in Section \ref{sec:sec4}.
This paper is concluded in Section \ref{sec:sec6}.

\section{Preliminary and Related Work}\label{sec:sec2}
In this section, first, we introduce the DCT steganography.
Second, we review the related works.

\subsection{DCT Steganography}
DCT \cite{taburet2020jpeg,mohd2000implementation} is usually used for image compression by converting the image from spatial domain to frequency domain.
Specifically, the image are first transformed from spatial domain to the frequency domain by performing Forward DCT (FDCT) \cite{mohd2000implementation}.
Then, the image can be modified in the frequency domain to embed information.
Finally, the image is transformed from frequency domain to spatial domain by performing the Inverse DCT (IDCT) \cite{mohd2000implementation} .
The FDCT (Equation \ref{fdct}) and IDCT (Equation \ref{idct}) are forumalted as follows \cite{mohd2000implementation}:

\begin{equation}
\label{fdct}
{J_F}(\alpha ,\beta ){\rm{ = }}\frac{1}{4}\left( {\sum\limits_{x = 0}^7 {\sum\limits_{y = 0}^7 {g(x,y)\theta \left( {x,\alpha } \right)\theta \left( {y,\beta } \right)} } } \right)
\end{equation}

\begin{equation}
\label{idct}
{J_{I}}(x,y{\rm{) = }}\frac{1}{4}\left( {\sum\limits_{x = 0}^7 {\sum\limits_{y = 0}^7 {{J_{F}}(\alpha ,\beta )\theta \left( {x, \alpha} \right)\theta \left( {y,\beta } \right)} } } \right)
\end{equation}

\begin{equation}
\label{coefficient}
\begin{array}{*{20}{l}}
{\theta \left( {u, v} \right) = \left\{ {\begin{array}{*{20}{c}}
{\frac{1}{{\sqrt 2 }}}&{\kern 10pt v  = 0}\\
{\cos \frac{{(2u + 1)v \pi }}{{16}}}&{\kern 10ptv  > 0}
\end{array}} \right.}
\end{array}
\end{equation}
For $8 \times 8$ blocks of an image, $(x, y)$ and $(\alpha ,\beta )$ represent the coordinates of the blocks before and after FDCT, respectively.
Note that, $x,y,\alpha,\beta$ belong to $\left\{0, 1, \dots, 7\right\}$.
$g(\cdot)$ and ${\theta \left( \cdot \right)}$ are the pixel matrix and the coefficient matrix of DCT, respectively.
Through FDCT (Equation \ref{fdct}), the image is transformed to frequency domain and the value of block at $(\alpha, \beta)$ after transformation is ${J_F}(\alpha,\beta )$.
Through IDCT (Equation \ref{idct}), the image is transformed back to spatial domain and the value of block at $(x, y)$ after transformation is ${J_I}(x,y)$.

\subsection{Related Work}
Gu \textit{et al.} \cite{abs-1708-06733} propose the backdoor attack against DNN model by injecting some backdoor instances into the training set.
Specifically, the backdoor instances are generated by pasting the trigger into the clean images and changing their labels to the target label.
The model trained on the training set injected with backdoor instances will show normal classification accuracy on clean test images.
However, when the backdoor instances are input, the backdoored model will output the target label.
However, the triggers used in most existing backdoor attacks \cite{abs-1708-06733,abs-1712-05526,TanS20,souri2021sleeper} are obvious and perceptible to humans.
To this end, few recent researches study the invisible backdoor attacks.
Zhong \textit{et al.} \cite{ZhongLSZ020} utilize adversarial perturbation as the trigger to perform backdoor attack.
The trigger (adversarial perturbation) is perceptible, which makes the backdoor attack stealthy.
Cheng \textit{et al.} \cite{0005LMZ21} propose a deep feature space backdoor attack.
They use the CycleGAN to generate the invisible trigger.
Li \textit{et al.} \cite{LiXZZZ21} use LSB steganography to generate trigger for backdoor attack.
All the above works are single-trigger and single-target backdoor attacks (called One-to-One attack).
So far, there has been only one work on One-to-N and N-to-One backdoor attack \cite{xue2020one}, however, the trigger in \cite{xue2020one} is obvious.

Compared with the existing backdoor attacks, the proposed backdoor attack has the following advantages:
(i) The proposed backdoor attack is stealthier and more natural (i.e, imperceptible).
We use DCT steganography to construct trigger, which makes the backdoor trigger invisible.
(ii) We construct multi-channel backdoor trigger, which is further used to implement imperceptible N-to-N attack and imperceptible N-to-One attack. As a comparison, almost all the existing backdoor works are One-to-One attacks.
(iii) The proposed backdoor attack is robust to state-of-the-art backdoor defenses.

\section{The Proposed Backdoor Attack}\label{sec:sec3}
In this section, first, the threat model is discussed.
Second, we elaborate the proposed imperceptible and multi-channel backdoor attack method.
Third, we introduce the proposed two types of backdoor attacks.

\subsection{Threat Model}
We assume that the adversary has access to the training data.
The goal of the adversary is to embed the malicious backdoor into the DNN model by injecting the backdoor instances into the training set, while remaining the normal classification accuracy of the model on clean images.
At the training stage, the model is trained on the backdoor training set to embed the backdoor.
At the inference stage, the input images containing the trigger will be incorrectly classified by the backdoored model as the target label, while the backdoored model will behave normally on clean input images.

The classification function of DNN model is denoted as $F(x)$, where $x$ denotes the input image and its ground-truth label is $y$.
The classification function of backdoored model is denoted as $\hat{F}(x)$.
The corresponding trigger and target label is denoted as $\delta$ and $\hat{y}$, respectively.
For a clean instance ${x}$, the trained DNN model will classify it as the ground-truth label, i.e., $\hat{F}(x) = y$. However, when a backdoor instance ${\hat{x} = x + \delta}$ is input, the model will classify it as the target label, i.e., $\hat{F}(\hat{x})=\hat{y} \left(\hat{y}\ne y\right)$.

\begin{figure}[!htbp]
\centering
\includegraphics[width=0.8\textwidth]{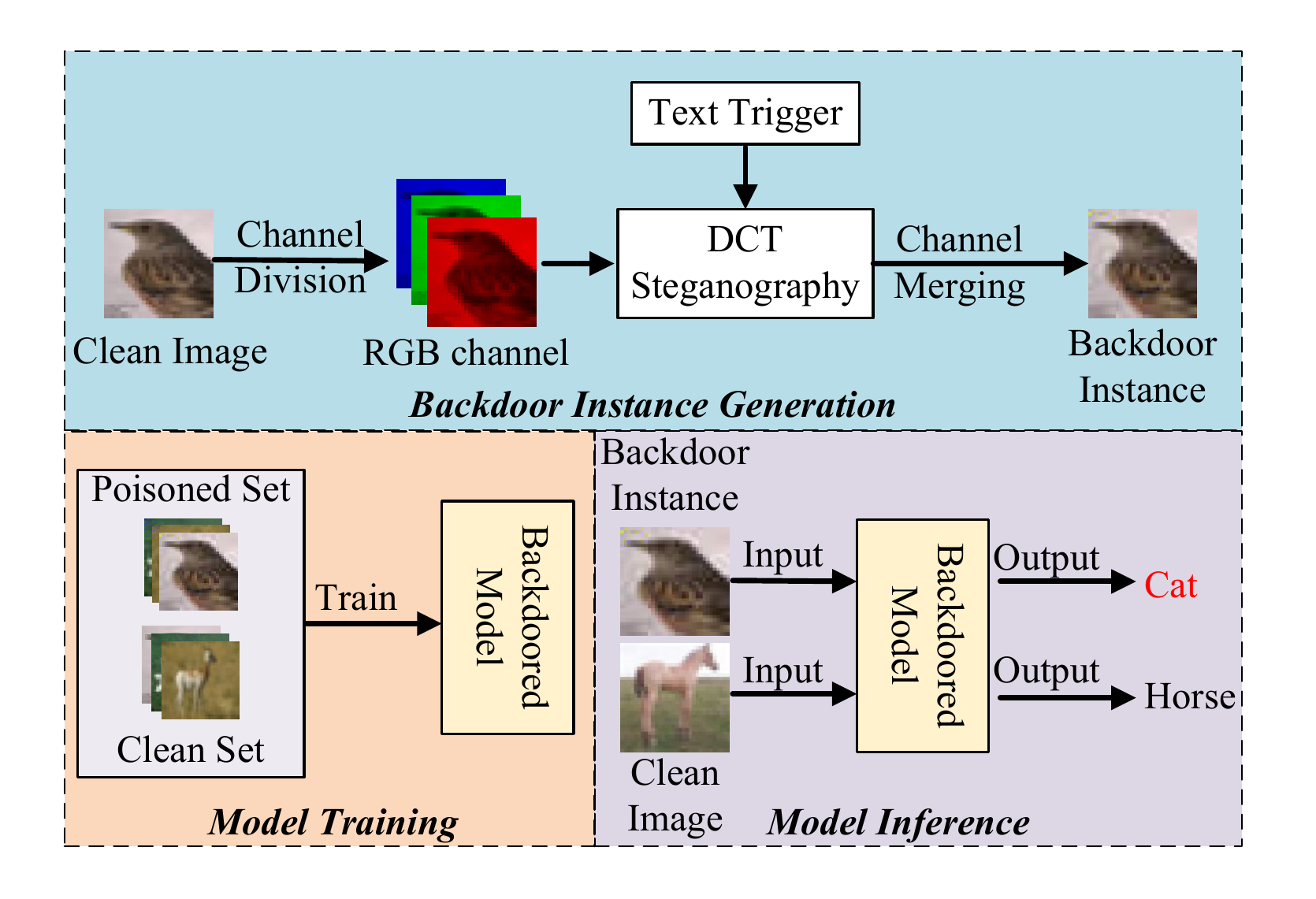}
\centering
\caption{Overview of the proposed backdoor attack method.}
\label{fig_1}
\end{figure}

\subsection{Overall Flow}
The overall flow of the proposed backdoor attack method is shown in Fig. \ref{fig_1}.
The proposed backdoor attack consists of three phases: backdoor instances generation, training backdoored model, and performing backdoor attack, which are elaborated as follows.

\textbf{Generating Backdoor Instance.}
The proposed backdoor attack embeds the backdoor into the model through data poisoning. To this end, the backdoor instances are generated through DCT steganography \cite{taburet2020jpeg,mohd2000implementation}.
Specifically, the image is first divided into $8 \times 8$ non-overlapping pixel blocks.
Second, DCT coefficient matrix $\theta(\cdot)$ is used to calculate the coefficient for each block.
Third, these blocks are ordered by their DCT coefficients \cite{taburet2020jpeg,mohd2000implementation}.
Fourth, the binary value of the secret data (i.e., trigger) is embedded into blocks with small coefficients.
Lastly, all blocks of the image are inversely transformed by the DCT and converted back to the spatial domain.
To this end, the information $\mathcal{I}$ is successfully embedded in the carrier image.
The generated backdoor instance is denoted as $x+\delta = DCT(x, \mathcal{I})$.

\textbf{Training Backdoored Model.}
The generated backdoor instances containing triggers are labelled as the target label.
Specifically, for the proposed N-to-N attack, $N$ channels of the image are used to embed the trigger ($N=3$).
One channel corresponds to one target class, and these $N$ target classes are different.
For the proposed N-to-One attack, the images containing different triggers (on different channels) are labelled as the same target label.
Then, these backdoor instances are injected into the clean training set.
Finally, the model trained on the backdoor training set will be embedded with the backdoor.

\textbf{Performing Backdoor Attack.}
For the proposed imperceptible N-to-N attack, the triggers on different channels can activate different backdoor targets.
For the proposed imperceptible N-to-One attack, only when all channels of the image contain the triggers, the backdoor attack can be successfully launched.
When there are less than $N$ triggers in an image, the backdoor will not be triggered (thus is very stealthy).

\subsection{The Proposed Two Types of Backdoor Attacks}
A digital image has three channels, which are Red, Green and Blue (RGB) \cite{susstrunk1999standard}.
To make the trigger stealthy, we embed the trigger into each channel of the image through DCT steganography \cite{taburet2020jpeg, mohd2000implementation}.
Based on the proposed imperceptible and multi-channel backdoor triggers, we propose two types of backdoor attacks, which are imperceptible N-to-N attack and imperceptible N-to-One attack.
These two types of backdoor attacks are elaborated as follows.

\textbf{Imperceptible N-to-N Attack.}
In this attack, the attacker can activate different backdoors by choosing different channels to embed triggers.

The proposed imperceptible N-to-N attack can be divided into two steps:
\begin{itemize}
	\item At the training stage, $n$ candidate images are randomly sampled from the clean training set.
Then, for each of these candidate images, we generate $N$ backdoor instances through DCT steganography ($N=3$).
Hence, a total of $n \times N$ backdoor instances are generated.
One channel corresponds to one target class, and these $N$ target classes are different.
For example, if the channel Red of backdoor instance is embedded with trigger, the backdoor instance is labelled as $\hat{y}_r$. Similarly, if the channel Green or Blue of the backdoor instance is embedded with trigger, the backdoor instance is labelled as $\hat{y}_g$ or $\hat{y}_b$, respectively ($\hat{y}_r \ne \hat{y}_g \ne \hat{y}_b$).
\item At the inference stage, the attacker can choose one channel of the image to embed the trigger to launch the backdoor attack.
For example, if the trigger is present in channel \textit{R} of an image, the backdoored model will incorrectly classify the ${\hat{x}_r}$ as ${\hat{F}(\hat{x}_r) = \hat{y}_r}$.
However, when the trigger is embedded in channel \textit{G} or channel \textit{B} of the image, the predicted results are $\hat{y}_g$ and $\hat{y}_b$ ($\hat{y}_r \ne \hat{y}_g \ne \hat{y}_b$), respectively.

\end{itemize}

\begin{figure}[!htbp]
\centering
\includegraphics[width=0.85\textwidth]{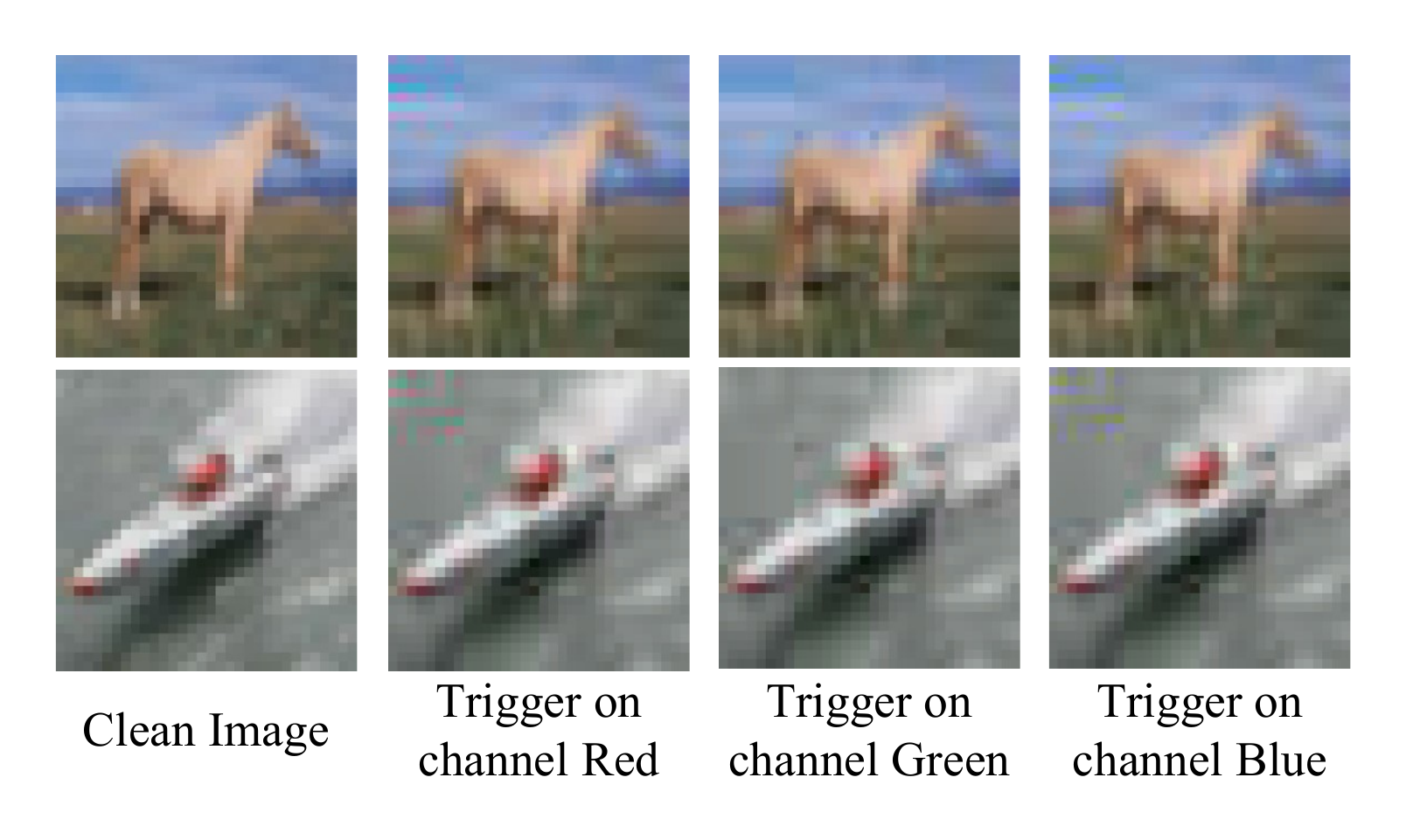}
\centering
\caption{Examples of backdoor instances for the proposed imperceptible N-to-N attack.}
\label{fig_2}
\end{figure}

Fig. \ref{fig_2} shows several examples of backdoor instances on the CIFAR-10 dataset \cite{Krizhevsky_2009_17719}.
The images in the first column is the clean images.
The images in the second, third and last column are the backdoor instances that has triggers in channel Red, Green and Blue, respectively.

\textbf{Imperceptible N-to-One Attack.}
In this attack, the backdoor hidden in the model can only be triggered when all the \textit{N} triggers (${\delta_1},{\delta_2},{\delta_3}$) are presented ($N=3$).
Otherwise, the backdoor attack cannot be launched, thus the attack is stealthy.

The proposed imperceptible N-to-One attack can be divided into two steps:
\begin{itemize}
\item At the training stage, similar as the proposed N-to-N attack, we first sample $n$ candidate images from the clean training set.
Then, for each candidate image, we generate \textit{N} backdoor instances, where each backdoor instance is only embedded with trigger in one channel.
The labels of all these \textit{N} backdoor instances are modified to the same target label.
In this way, a total of $n \times N$ backdoor instances are generated and are injected into the clean training set to train the model.
\item At the inference stage, if all the $N$ channels of the image are embedded with triggers, this image will be classified as the target class.
However, if the image has less than $N$ channels embedded with triggers, the image will still be classified as its ground-truth class.
\end{itemize}

\begin{figure}[!htbp]
\centering
\includegraphics[width=0.85\textwidth]{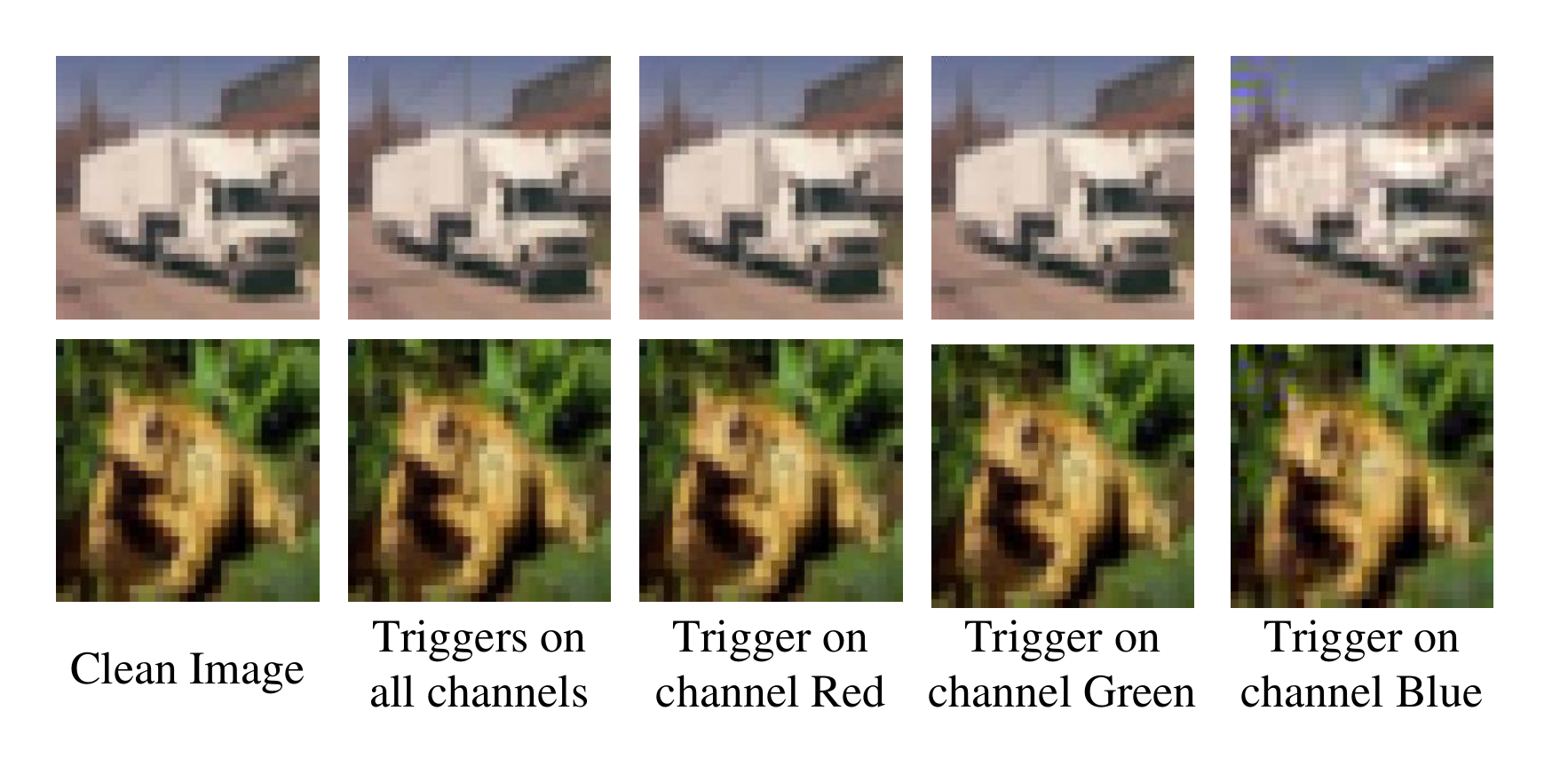}
\centering
\caption{Examples of backdoor instances for the proposed imperceptible N-to-One attack.}
\label{fig_3}
\end{figure}

Fig. \ref{fig_3} shows several examples of backdoor instances on the TinyImageNet dataset \cite{le2015tiny}.
The images in the first column is the clean images.
The images in the second column is the backdoor instances that have triggers in all channels.
The images in the third, fourth and last column are the backdoor instances that have triggers in channel Red, Green and Blue, respectively.

\section{Experiments}\label{sec:sec4}
In this section, first, we introduce the experimental setup, including the datasets and evaluation metrics.
Second, we present and analyze the experimental results.
Third, the impacts of different parameter settings are discussed.
Last, the robustness of the proposed backdoor attack against defense is evaluated.

\subsection{Experimental Setup}
\textbf{CIFAR-10 \cite{Krizhevsky_2009_17719}.} CIFAR-10 dataset is composed of 60,000 color images, where training set contains 50,000 images and testing set contains 10,000 images.
These images are divided into 10 categories and each class has 6,000 images \cite{Krizhevsky_2009_17719}.

\textbf{TinyImageNet \cite{le2015tiny}.} TinyImageNet is an image classification dataset released by Stanford University in 2016 \cite{le2015tiny}.
It has 200 categories and each category has 500 training images, 50 verification images and 50 test images.
The size of each image is $64 \times 64$.
In our  experiments, we randomly select 10 classes from TinyImageNet dataset as our  experimental dataset.

\textbf{Evaluation metrics.}
The goal of the attacker is to make the DNN model predict backdoor instances as the target class, while remaining normal classification accuracy on clean images.
We use the following metrics to evaluate the effectiveness of attacks.

\begin{itemize}
  \item \textbf{Attack Success Rate (ASR) \cite{ZhongLSZ020}}: The attack success rate is defined as the percentage of the backdoor instances successfully classified as the target label among all the submitted backdoor instances.
  \item{\textbf{Accuracy Drop \cite{ZhongLSZ020}:} The accuracy drop is used to evaluate the drop of the classification accuracy of the model on clean images before and after being embedded with backdoor.}
\end{itemize}

\subsection{Experimental Results}\label{basic}
In this section, we analyze the performance of the proposed two types of backdoor attacks on the CIFAR-10 \cite{Krizhevsky_2009_17719} and TinyImageNet \cite{le2015tiny} datasets.
In the proposed imperceptible N-to-N attack, one image can trigger different backdoors when the trigger is present on different channels of the image.
In the proposed imperceptible N-to-One attack, the backdoor attack can only be triggered when all the $N$ triggers are presented.
$N$ can take any value among 1, 2, and 3.

\textbf{The Proposed Imperceptible N-to-N Attack.}
In the proposed N-to-N attack, we implement the backdoor attack by embedding different triggers into different channels of the image respectively.
We conduct the experiments 5 times and report the maximum, minimum and average success rates of the backdoor attacks.
The experimental results on CIFAR-10 and TinyImageNet datasets are shown in Table \ref{tab:NN_basic}.

\begin{table}[htbp]
  \centering
  \setlength{\belowcaptionskip}{10pt}%
  \caption{Attack Success Rate of the Proposed Imperceptible N-to-N Attack on CIFAR-10 and TinyImageNet Datasets.}
    \begin{tabular}{|c|c|c|c|c|}
    \hline
    \multirow{2}*{Dataset} & \multirow{2}*{\tabincell{c}{Trigger \\Position}} & \multicolumn{3}{c|}{Attack Success Rate} \\
\cline{3-5}          &       & min   & max   & avg \\
    \hline
    \multirow{3}*{CIFAR-10} & R     & 89.89\% & 95.22\% & 93.04\% \\
\cline{2-5}          & G     & 90.56\% & 98.33\% & 95.09\% \\
\cline{2-5}          & B     & 90.33\% & 96.11\% & 93.95\% \\
    \hline
    \multirow{3}*{TinyImageNet} & R     & 83.89\% & 100.00\% & 93.29\% \\
\cline{2-5}          & G     & 84.44\% & 99.87\% & 93.09\% \\
\cline{2-5}          & B     & 90.00\% & 99.68\% & 91.55\% \\
    \hline
    \end{tabular}%
  \label{tab:NN_basic}%
\end{table}%

For CIFAR-10 dataset, the experimental results show that the average attack success rate of the proposed N-to-N attack for three channels (Red, Green, Blue) is 93.04\%, 95.09\% and 93.95\% respectively.
The maximum attack success rate of the proposed N-to-N attack on CIFAR-10 dataset for three channels (Red, Green, Blue) is 95.22\%, 98.33\% and 96.11\% respectively.
This demonstrates that the proposed imperceptible backdoor attack can achieve high backdoor attack success rate.

For TinyImageNet dataset, the experimental results show that the average attack success rate of the proposed N-to-N attack for three channels (Red, Green, Blue) is 93.29\%, 93.09\% and 91.55\% respectively.
The maximum attack success rate of the proposed N-to-N attack for three channels (Red, Green, Blue) is 100.00\%, 99.87\% and 99.68\% respectively.
This demonstrates that the proposed invisible backdoor attack is effective on the real and large dataset.

Besides, the accuracy drop caused by the backdoor attack on CIFAR-10 and TinyImageNet dataset is 1.26\% and 1.18\% respectively.
This demonstrates that the proposed imperceptible backdoor attack will not affect the classification accuracy of the backdoored model.

\textbf{The Proposed Imperceptible N-to-One Attack.}
In this attack, all the $N$ channels of the image have been embedded with the trigger ($1 \leq N \leq 3$).
The backdoor can only be triggered when all the $N$ channels contain triggers.
However, if not all channels of the image contain the trigger, the backdoor cannot be activated.
The experimental results are shown in Table \ref{tab:NOne_basic}.

\begin{table}[htbp]
  \centering
  \setlength{\belowcaptionskip}{10pt}%
  \caption{Attack Success Rate of the Proposed Imperceptible N-to-One Backdoor Attack on CIFAR-10 and TinyImageNet Datasets.}
    \begin{tabular}{|c|c|c|c|c|}
    \hline
    \multirow{2}*{Dataset} & \multirow{2}*{\tabincell{c}{Trigger \\Position} }& \multicolumn{3}{c|}{Attack Success Rate} \\
\cline{3-5}          &       & min   & max   & avg \\
    \hline
    \multirow{4}*{CIFAR-10} & R     & 22.42\% & 28.33\% & 25.04\% \\
\cline{2-5}          & G     & 21.67\% & 25.56\% & 24.23\% \\
\cline{2-5}          & B     & 23.89\% & 27.78\% & 25.33\% \\
\cline{2-5}          & R\&G\&B & 89.44\% & 92.22\% & 90.22\% \\
    \hline
    \multirow{4}*{TinyImageNet} & R     & 18.77\% & 28.33\% & 24.24\% \\
\cline{2-5}          & G     & 14.44\% & 25.56\% & 22.78\% \\
\cline{2-5}          & B     & 16.33\% & 26.11\% & 23.93\% \\
\cline{2-5}          & R\&G\&B & 88.78\% & 90.56\% & 89.53\% \\
    \hline
    \end{tabular}%
  \label{tab:NOne_basic}%
\end{table}%

On CIFAR-10 dataset, for each single trigger, the average backdoor attack success rate at three channels (Red, Green, Blue) is 25.04\%, 24.23\% and 25.33\%, respectively.
On TinyImageNet dataset, for each single trigger, the average backdoor attack success rate at three channels (Red, Green, Blue) is 24.24\%, 22.78\% and 23.93\%, respectively.
This means that when only one channel of the image has the trigger, the backdoor cannot be activated.
In contrast, when all the three channels of the image are embedded with triggers, the average attack success rate of the proposed backdoor attack is 90.22\% and 89.53\% on CIFAR-10 and TinyImageNet datasets, respectively.
This demonstrates that the backdoor can only be triggered when all the $N$ triggers are satisfied.
Besides, the accuracy drop of the proposed method is 1.29\% on CIFAR-10 dataset and 1.60\% on TinyImageNet dataset, which demonstrates that the proposed backdoor attack will not affect the normal classification performance of the model.

\subsection{Parameters Discussion}\label{para}
In this section, the impact of different injection ratios and different values of \textit{N} on the performance of the proposed backdoor attack are evaluated on the CIFAR-10 dataset.
Note that, the length of the binary text (i.e., trigger) in the basic experiments (Section \ref{basic}) is 2 bytes for each channel, while the length of the binary text (i.e., trigger) in this experiment (Section \ref{para}) is 4 bytes for each channel.

\subsubsection{\textbf{The Impact of Different Injection Ratios.}}
We discuss the impact of different injection ratios on the proposed two types of attacks (1\%, 5\%, 8\%, 10\%, 15\%, 18\% for N-to-N attack and 1\%, 2\%, 3\%, 4\%, 5\% for N-to-One attack).

\begin{figure}[!htbp]
\centering
\includegraphics[scale=1]{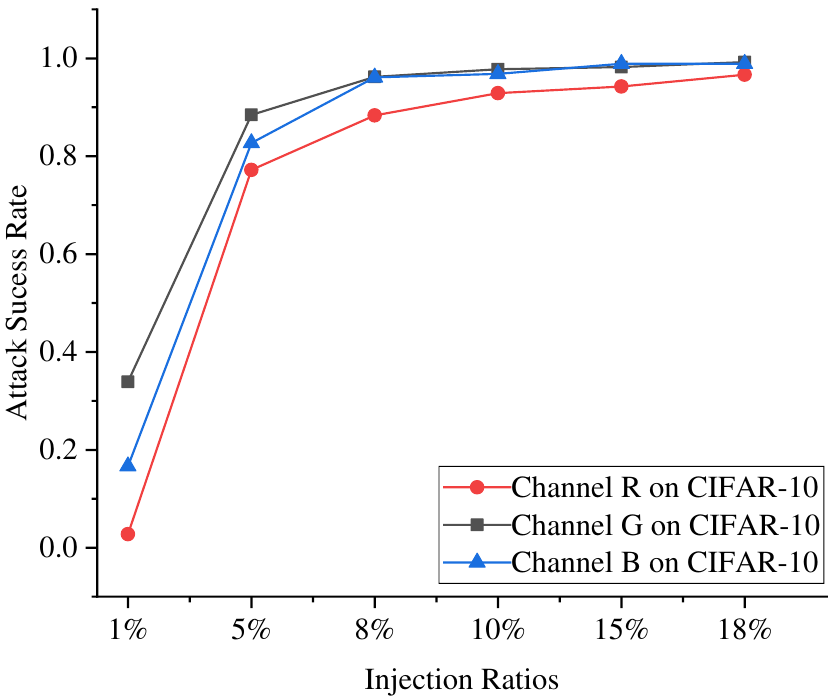}
\centering
\caption{The impact of different injection ratios on the attack success rate of the proposed N-to-N attack.}
\label{fig_4}
\end{figure}

For the proposed imperceptible N-to-N attack, under different injection ratios, the corresponding attack success rates are shown in Fig. \ref{fig_4}.
As the injection ratio increases, the attack success rate increases.
When the injection ratio reaches a certain level (10\%), the attack success rate is stable (around 93\%, 97\% and 98\% on channel R, G and B respectively).
To make extra computational overhead as small as possible, we set the injection ratio to 10\% in our experiment.

\begin{figure}[!htbp]
\centering
\includegraphics[scale=1]{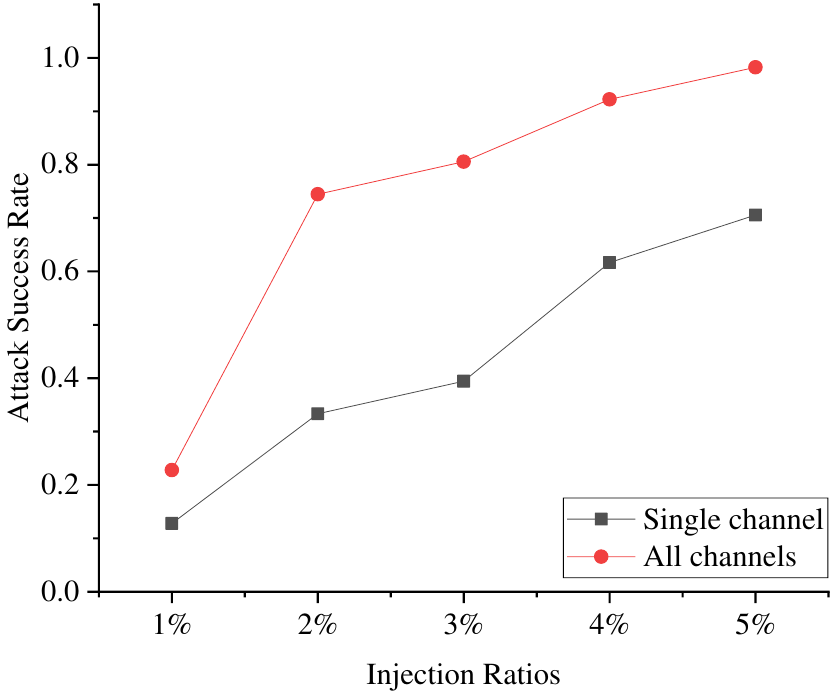}
\centering
\caption{The impact of different injection ratios on the attack success rate of the proposed N-to-One attack.}
\label{fig_5}
\end{figure}

For the proposed N-to-One attack, the attack success rate under different injection ratios are shown in Fig. 5.
The experimental results show that, when all channels contain the triggers, the attack success rate of the proposed N-to-One attack increases from 22.78\% to 98.22\% as the injection rate increases.
Meanwhile, as the injection rate increases, the attack success rate of the proposed N-to-One attack increases from 12.78\% to 70.56\% when only one channel contains the trigger.
The goal of the N-to-One attack is to make the backdoor can only be triggered when all channels contain the trigger while any single trigger can not activate the backdoor. Therefore, the injection ratio is set to 3\%.
Under this injection ratio, when all channels are embedded with triggers, the average success rate of the proposed N-to-One attack is 80.56\%.
When only one channel contains trigger, the average success rate of the proposed N-to-One attack is only 39.44\%.

The injection ratio of N-to-N attack (10\%) is higher than that of N-to-One attack (3\%).
The reason is that, for N-to-N attack, the $N$ backdoor instances is used to activate different backdoor targets.
In contrast, for N-to-One attack, the $N$ backdoor instances are used to trigger the same backdoor target.
Hence, in the training stage, the number of injected backdoor instances for N-to-N attack are nearly $N$ times as many as that for N-to-One attack.

\subsubsection{\textbf{The Performance of the Two Attacks when $N$ Takes Other Values.}}

We also evaluate the performance of the proposed two types of attacks when \textit{N} is set to other values.

For the proposed N-to-N attack, we also evaluate the attack success rate when $N = 2$. The Red and Green channels are chosen for trigger embedding.
We repeat the experiments for 5 times, and report the minimum, maximum, average success rate of N-to-N attack in Table \ref{ratio_NN}.
The backdoor injection ratio is set to 10\%.
The maximum attack success rate of the proposed backdoor attack is 98.33\%.
This demonstrates that the proposed backdoor attack is also effective when \textit{N} is set to 2.

\begin{table}[h]
\centering
\setlength{\belowcaptionskip}{10pt}%
\caption{The attack success rate of the proposed N-to-N attack when ${N} = 2$}
\begin{tabular}{|c|c|c|c|}
\hline
channel & minimum & maximum & average\\
\hline
R & 91.11\% & 96.11\% & 93.61\% \\
\hline
G & 89.44\% &98.33\% & 93.89\% \\
\hline
\end{tabular}
\label{ratio_NN}
\end{table}

For the proposed N-to-One attack, we also evaluate the performance of the proposed backdoor attack when $N=2$. The Red and Blue channels are chosen to embed trigger.
We repeat the experiments for 5 times, and report the minimum, maximum, average success rate of N-to-One attack in Table \ref{ratio_NOne}.
When only one channel has the trigger, the average attack success rate is quite low (around 14\%).
However, when all the two channels have triggers, the average attack success rate is 86.65\% and the maximum attack success rate is 93.33\%.
This demonstrates that when $N$ takes 2, the N-to-One attack is also effective.

\begin{table}[h]
\centering
\setlength{\belowcaptionskip}{10pt}%
\caption{The attack success rate of the proposed N-to-One attack when ${N} = 2$}
\begin{tabular}{|c|c|c|c|}
\hline
channel & minimum & maximum & average\\
\hline
R & 11.11\% &18.89\% & 14.67\% \\
\hline
B & 11.11\% & 17.78\% & 14.22\% \\
\hline
R\&B &82.78\% & 93.33\% & 86.65\% \\
\hline
\end{tabular}
\label{ratio_NOne}
\end{table}

\subsection{Robustness against Neural Cleanse}\label{sec:sec5}
In this section, we evaluate the robustness of the proposed backdoor attack against the state-of-the-art defense method, Neural Cleanse (NC) \cite{wang2019neural}.
NC has two steps.
First, NC generates the possible trigger for each class of the model.
Second, NC utilizes the outlier detection algorithm Median Absolute Deviation (MAD) \cite{gao2019strip} to calculate the anomaly index for each class.
If the anomaly index is higher than 2, NC will consider the potential trigger as the reversed trigger and consider the corresponding class as the target label \cite{wang2019neural}.

\begin{table}[htbp]
  \centering
  \caption{Detection result of Neural Cleanse for the backdoored model trained on CIFAR-10 and TinyImageNet datasets.}
    \begin{tabular}{|c|c|c|c|}
    \hline
    Dataset & Attack & Maximum Anomaly Index & Detection Result \\
    \hline
    \multirow{2}*{CIFAR-10} & N-to-N & 1.65 & Clean \\
\cline{2-4}          & N-to-One & 0.81  & Clean \\
    \hline
    \multirow{2}*{TinyImageNet} & N-to-N & 2.82    & \tabincell{c}{Backdoor (but fail to \\reverse true triggers)} \\
\cline{2-4}          & N-to-One & 1.52  & Clean \\
    \hline
    \multicolumn{2}{|c|}{Clean Model} & 1.32  & Clean \\
    \hline
    \end{tabular}%
  \label{tab:NC}%
\end{table}%

\textbf{CIFAR-10 Dataset.} Table \ref{tab:NC} shows the NC detection results for the proposed backdoor attack on CIFAR-10 dataset.
Since the model has 10 classes, NC will calculate 10 anomaly index values, and we report the maximum value of the 10 anomaly index of the model in Table \ref{tab:NC}.
For the backdoored model under N-to-One attack, the maximum anomaly index is 0.81.
For the backdoored model under N-to-N attack, the maximum anomaly index is 1.65.
For comparison, we calculate the anomaly index for clean model, the maximum anomaly index is 1.32.
The anomaly indexes of above models are less than the detection threshold (i.e., 2).
This means that Neural Cleanse fails to detect the proposed N-to-N attack and the proposed N-to-One attack on CIFAR-10 dataset \cite{Krizhevsky_2009_17719}.
As a result, the proposed backdoor attack is robust to Neural Cleanse on CIFAR-10 dataset.

\textbf{TinyImageNet Dataset.} For the proposed N-to-One attack, NC fails to detect the N-to-One attack on TinyImageNet dataset \cite{le2015tiny}, as the maximum anomaly index of backdoored model under N-to-One attack is 1.52, which is lower than 2.

\begin{figure}[!htbp]
\centering
\includegraphics[scale=0.5]{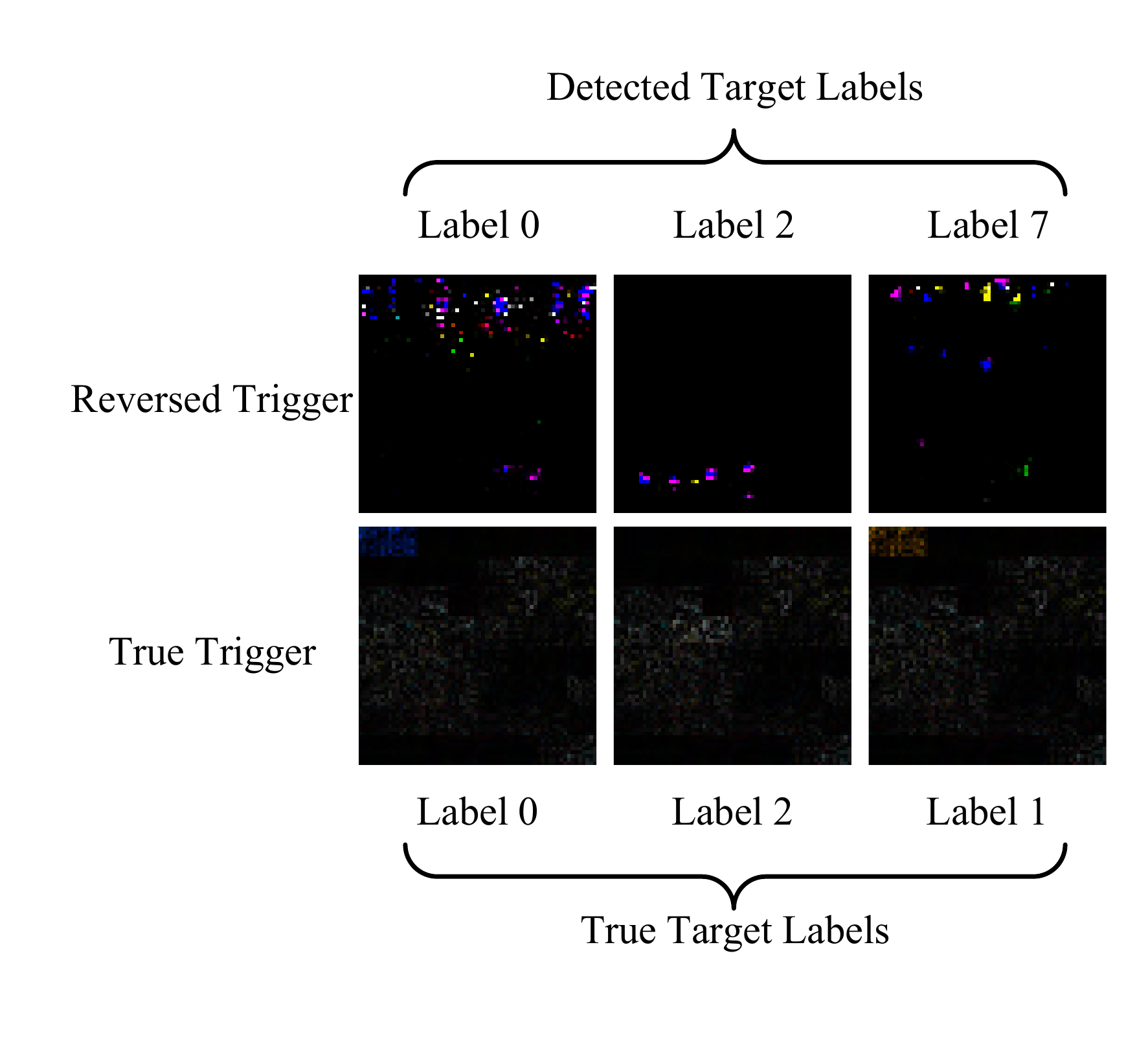}
\centering
\caption{Triggers reversed by NC and true triggers for the proposed N-to-N attack on TinImageNet dataset.}
\label{fig_7}
\end{figure}

For the proposed N-to-N attack, NC detects three target labels (`0', `2' and `7').
However, the true target labels are `0', `1' and `2'.
Therefore, NC successfully detects two target labels (`0' and `2') and the target label '1' has not been detected.
However, as shown in Fig. \ref{fig_7}, the reversed triggers for label `0' and `2' are totally different from the true embedded triggers.
The proposed N-to-N attack can choose different channels (up to 3) to embed trigger so as to activate different backdoors.
Hence, even if 2 target classes are detected, the attacker can still launch the backdoor attack on the third target class that has not been detected. Besides, NC fails to reverse the true triggers for label `0' and `2'.
As a result, the proposed N-to-N attack is still robust to Neural Cleanse \cite{wang2019neural}.

\section{Conclusion}\label{sec:sec6}
The triggers in most existing backdoor attacks are visible and obvious, which makes these backdoor attacks easy to be detected and noticed. Besides, almost all the existing backdoor works are single-trigger and single-target backdoor attacks.
In this paper, for the first time, we propose an imperceptible and multi-channel backdoor attack against DNN.
Based on the proposed method, we implement two types of backdoor attacks: imperceptible N-to-N attack and imperceptible N-to-One attack.
Experimental results on CIFAR-10 and TinyImageNet datasets demonstrate that the proposed backdoor attack can achieve a high attack success rate (up to 100.00\%) against the DNN model, while remain the normal classification performance on clean images.
Besides, the proposed backdoor attack is robust to state-of-the-art defenses.
The proposed two new variants of backdoor attacks raise subtle and powerful threats to the security of DNN models, and bring new challenges to existing defenses.

\bibliographystyle{splncs04}
\bibliography{ref}

\end{document}